\documentclass{Interspeech2024}

\interspeechcameraready 

\usepackage{multirow}

\title{A Small and Fast BERT for Chinese Medical Punctuation Restoration}

\name[affiliation={1,2}]{Tongtao}{Ling}
\name[affiliation={1}]{Yutao}{Lai}
\name[affiliation={1}]{Lei}{Chen}
\name[affiliation={3}]{Shilei}{Huang}
\name[affiliation={2,*}]{Yi}{Liu}

\address{
  $^1$Guangdong University of Technology, Guangzhou, China\\
  $^2$IMSL Shenzhen Key Lab, PKU-HKUST Shenzhen Hong Kong Institute, Shenzhen, China \\
  $^3$Shenzhen Raisound Technologies, Co., Ltd, Shenzhen, China
  {\thanks{* corresponding author}}}
\email{ltt\_rick@163.com, 2112114007@mail2.gdut.edu.cn, chenlei3@gudt.edu.cn, shilei.huang@raisound.com, yi.liu@imsl.org.cn}


\begin{document}

\maketitle

    
    
    

\begin{abstract}

In clinical dictation, utterances after automatic speech recognition (ASR) without explicit punctuation marks may lead to the misunderstanding of dictated reports. To provide a precise and understandable clinical report with ASR, automatic punctuation restoration (APR) is required. Considering a practical scenario, we propose a fast and lightweight pre-trained model for Chinese medical punctuation restoration based on the `pre-training and fine-tuning' paradigm. In this work, we distill pre-trained models by incorporating supervised contrastive learning and a novel auxiliary pre-training task (Punctuation Mark Prediction) to make it well-suited for punctuation restoration. We then reformulate APR as a slot tagging problem in the fine-tuning stage to bridge the gap between pre-training and fine-tuning. Our experiments on various distilled models reveal that our model can achieve 95\% performance with a 10\% model size relative to the state-of-the-art Chinese RoBERTa. 

\end{abstract}

\section{Introduction}

Automatic punctuation restoration is a crucial post-processing step of ASR systems. In the medical field, Electronic Medical Record systems (EMRs) leverage ASR technology to assist doctors in inputting medical reports rather than typing them manually, thereby enhancing the efficiency of medical professionals~\cite{hodgson2016risks,salloum2017deep}. However, the absence of punctuation marks in ASR-generated transcripts significantly impacts the comprehensiveness of medical reports. Consequently, it is non-trivial to predict and insert punctuation marks for ASR-generated transcripts or numerous text processing applications~\cite{fu2021improving}.

Over the past few years, the mainstream of punctuation restoration can be divided into three categories: 1) The first line~\cite{lin2020joint,huang2021token,shi2021incorporating} views punctuation restoration as a sequence labeling or token classification problem, which predicts whether each token is followed by a punctuation mark or not. 2) Second, some researchers~\cite{klejch2017sequence,wang2018self} treated it as a Seq2Seq task, which inputs a non-punctuation sequence and outputs a punctuated one. 3) The others~\cite{zhu2022unified,sunkara2020multimodal} propose a multi-modal framework to learn hybrid representations from audio and text samples for downstream classifiers to perform punctuation restoration.

Considering that text data is easier to access than audio data, BERT~\cite{devlin2019bert} and its variants~\cite{yao2023minirbt,cui2021pre} have been widely applied to enhance the effectiveness of token classification,  we focus on the first method. However, BERT was trained on large-scale punctuated dataset with punctuations but fine-tuned on non-punctuated dataset, which may critically damage the semantics in downstream task. Besides, the data imbalance problem of token classification should be addressed. For instance, in the benchmark dataset IWSLT2011~\cite{federico2012overview}, over 85\% of tokens are no-punctuation tokens. Concretely, we aim to train a small and fast model that can be applied to the Chinese medical field, demonstrating high performance.

Inspired by these observations, in this paper, we propose a novel auxiliary pre-training task to alleviate the damage of semantics and incorporate supervised contrastive learning (SCL)~\cite{khosla2020supervised} to solve the problem of data imbalance. To train a light model that has smaller hidden layer dimensions and transformer layers, we employ knowledge distillation (KD)~\cite{yang2020textbrewer} during the pre-training stage. In the fine-tuning stage, we re-formalize punctuation restoration as a Slot Tagging~\cite{kim2015weakly} task to narrow the gap between pre-training and fine-tuning. The main contributions of this paper:
\begin{itemize}
    \item We train a fast and lightweight Chinese BERT model by incorporating contrastive learning and knowledge distillation into our auxiliary pre-training task. This model can be effectively fine-tuned for Chinese medical punctuation restoration.
    \item Our experiments reveal that compared with SOTA Chinese RoBERTa, our model achieves its 95\% performance while only having its 10\% model size. 
    \item The ablation study reveals that our auxiliary task and supervised contrastive learning can enhance the performance of Chinese medical punctuation restoration.
\end{itemize}

\section{Background}
\label{sec:related}

\noindent\textbf{Punctuation Restoration.} In the early years, punctuation restoration models primarily relied on lexical and prosodic features, but the acoustic feature is often noisy and error-prone~\cite{szaszak2019leveraging}. Until the advent of transformer-based models, most of the punctuation restoration models are based on recurrent neural networks~\cite{xu2016investigating}. Recently, pre-trained language models can enhance the performance of downstream tasks (e.g., token classification), and punctuation restoration is often formulated as a token classification task~\cite{lai2023boosting}.

\noindent\textbf{Distillation.}
Knowledge distillation is a common method for compressing models, which aims to optimize a small student model to emulate a larger teacher model. As different layers of the transformer capture distinct information, the student model must acquire knowledge from both the final and intermediate layers of the teacher model~\cite{ashihara22_interspeech}. Knowledge distillation has shown to be effective for natural language processing and speech processing~\cite{sanh2019distilbert,chang2022distilhubert}. 


\noindent\textbf{Contrastive Learning.}
Contrastive learning focuses on training models by highlighting the difference between pairs of data instances~\cite{oord2018representation}. The core idea is to learn representations that make similar instances more alike while making dissimilar instances more distinct~\cite{khosla2020supervised}. 

\begin{figure}[t]
    \centering    
    \includegraphics[width=1.0\linewidth]{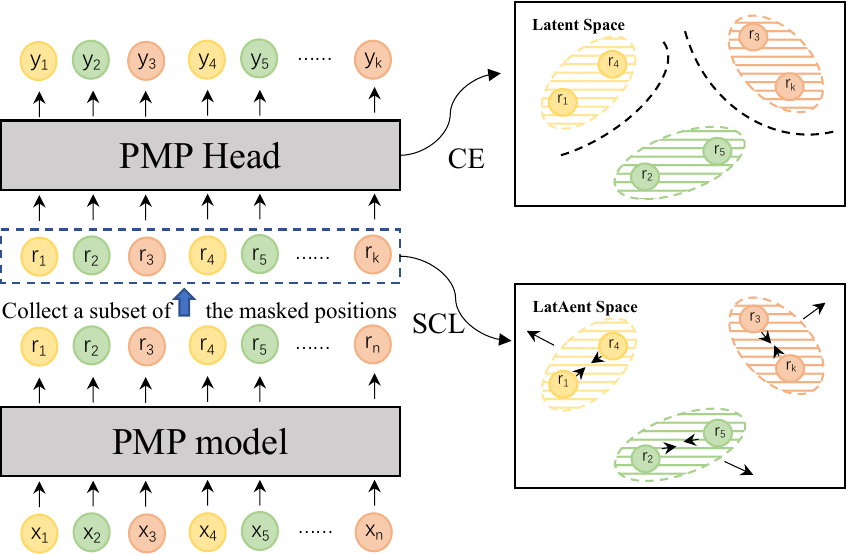}
    \caption{Overview of pre-training.}
    \label{fig:pretrain}
\end{figure}

\section{Approach}
\label{sec:appro}

\subsection{Punctuation Mark Prediction}

Masked Language Model (MLM)~\cite{devlin2019bert} is the main pre-training task in BERT and its variants, which learns contextual information that could be applied to downstream tasks (e.g., punctuation restoration). To boost the performance of punctuation prediction, we propose an auxiliary pre-training task called Punctuation Mark Prediction (PMP).

Specifically, as shown in Figure \ref{fig:pretrain}, given the input tokens $X = (x_1,...,x_n)$, PMP model computes the contextual representations $H^{(L)} = (r_1,...,r_n)$, $ H^{(L)} \in \mathbb{R}^{N \times d}$ by an embedding layer and a consecutive $L$-layer transformer, where $d$ is the dimension of hidden layer and $N$ is the maximum number of tokens.
\begin{equation}
    H^{(0)} = Embedding(X)
\end{equation}
\begin{equation}
    H^{(i)} = Transformer(H^{(i-1)}), i \in \{1,...,L\}
\end{equation}

PMP is similar to the MLM task, which aims to predict the token`s type at the {\tt [MASK]} position. After getting the contextual representations $H^{(L)}$, we collect a subset that contains each {\tt [MASK]} token representation $H^{(m)} \in \mathbb{R}^{k \times d} $, where $k$ is the number of {\tt [MASK]} tokens. Following original MLM, $k$ is set to $[N \times 15\%]$. Because the {\tt [MASK]} token doesn't exist in downstream tasks, therefore 80\% of the replaced tokens are {\tt [MASK]}, 10\% are random tokens, and 10\% are original tokens in the original MLM. Nevertheless, we utilize {\tt [MASK]} tokens in punctuation restoration and the masking probability is set to 100\%.
 
\begin{figure}[t]
    \centering    
    \includegraphics[width=1.0\linewidth]{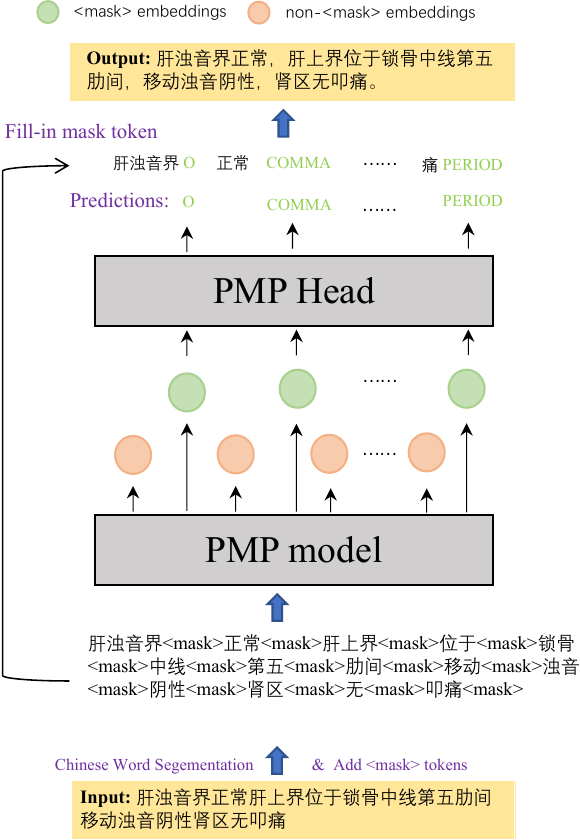}
    \caption{Overview of fine-tuning.}
    \label{fig:finetuning}
\end{figure}

To conduct punctuation mark prediction, PMP-Head is introduced to compute the probability $p$ over the punctuation mark set $Y$, such as \{\textit{O}, \textit{COMMA}(,), \textit{PERIOD}(.), \textit{QUESTION}(?)\}, \textit{O} denotes no-punctuation token. PMP-Head contains 
a weight matrix $W \in \mathbb{R}^{|Y| \times d}$ and a bias term $B \in \mathbb{R}^{|Y|}$. 

\begin{equation}
    p = H^{(m)}W^{\top} + B
\end{equation}

Considering the data imbalance of punctuation classes, we incorporate a supervised contrastive learning (SCL) loss with the standard cross-entropy (CE) loss. The core of contrastive learning is to compare one class with other classes in a batch, which represents that the same classes are pulled together in the latent space while simultaneously pushing apart from different classes, as shown in Fig \ref{fig:pretrain}. Since SCL is a supervised method, we pre-set a sample as anchor in each batch, the positive samples are defined as the ones in the sample class as the anchor, while the negative samples are at the opposite side. The CE and SCL loss are defined as:
\begin{equation}
    \mathcal{L}_{CE} = - \frac{1}{M}\sum^{M}_{i=1}y_i\log p_i
\end{equation}

\begin{equation}
    \mathcal{L}_{SCL} = \sum_{i \in I} \frac{-1}{|P(i)|} \sum_{p \in P(i)}\log \frac{\exp{(\Phi(r_i)\cdot \Phi(r_p)/\tau})}{\sum_{k\in A(i)}\exp{(\Phi(r_i)\cdot\Phi(r_k)/\tau)}}
\end{equation}
where $i$ is the anchor sample, $P(i)$ represents the set of positive samples and $A(i)$ represents the set of all samples, regardless of the sample type. $\Phi$ is $\ell_2$ normalization and $\tau$ is a scalar to stabilize the calculation. The contrastive loss is computed by similarity scores according to dot product. Finally, we add a hyper-parameter $\lambda$ to combine two losses.
\begin{equation}
    \mathcal{L}_{PMP} = (1-\lambda)\mathcal{L}_{CE} + \lambda\mathcal{L}_{SCL} 
\end{equation}

\begin{table*}[t]
    \caption{Comparison of model structures.} 
    \centering
    \label{tab:structure}
    \begin{tabular}{l|ccccc}
    \toprule
    Model & Layers & Hidden Size & FFN Size & MAHs & \#Params (Millions) \\
    \midrule
    RoBERTa-wwm (teacher)& 12 & 768 & 3072 & 12 & 102.3 (100\%) \\
    ALBERT-base-Chinese & 12 & 768& 3072& 12& 10.5 (10.3\%) \\
    RBT-6 & 6 & 768 & 3072 & 12 & 59.7 (58.4\%)\\
    MiniRBT & 6 & 256 & 1024 & 8 & 10.3 (10.1\%)\\
    TinyBERT & 4 & 312 & 1200 & 12 & 11.4 (11.2\%)\\
    \textbf{PMP-H768} & 6 & 768 & 3072 & 12 & 59.7 (58.4\%)\\
    \textbf{PMP-H256} & 6 & 256 & 1024 & 8 & 10.3 (10.1\%) \\
    \textbf{PMP-H312} & 4 & 312 & 1200 & 12 & 11.4 (11.2\%)\\
    \bottomrule
    \end{tabular}
\end{table*}

\begin{table}[t]
    \caption{Data distribution of CMRPT. }
    \label{tab:dataset}
    \resizebox{1\columnwidth}{!}{
    \begin{tabular}{cccccc}
    \toprule
    Dataset & Split & O (No punctuation) & Comma & Period & Colon\\
    \midrule
        \multirow{3}*{CMRPT}
        &Train&4,332,870&434,981&83,248&33,304\\
        &Dev&442,505&44,477&8,415&3,518\\
        &Test&44,834&4,283&786&334\\
    \bottomrule
    \end{tabular}
    }
\end{table}

\subsection{Knowledge Distillation}
During the stage of pre-training, we distill the PMP model that follows conventional knowledge distillation method (Teacher-Student). First, we utilize Chinese RoBERTa-wwm to initialize PMP model as our teacher model, and select several transformer layers from the teacher model to initialize our student model. Then we distill the hidden layer from teacher model to student model through the Mean Square Error (MSE) function. The objective is defined as:
\begin{equation}
    \mathcal{L}_{distill} = \frac{1}{M} \sum^{M}_{i=1} (H_{s}^{j^{'}}W_{h}-H_{t}^{j})^2
\end{equation}
where $H_{s}^{j^{'}}$ is the $j^{'}$-th layer of student model and $H_{t}^{j}$ is the $j$-th layer of teacher model. The weight matrix $W_h \in \mathbb{R}^{d^{'} \times d}$ matches the hidden layer of student model and the hidden layer of teacher model. $d^{'}$ and $d$ represent the dimension of hidden layer of student model and teacher model respectively.

\subsection{Slot Tagging for {\tt [MASK]}}

After distilling PMP model, vanilla MLM suffers the gap between pre-training stage and fine-tuning stage due to the absence of the artificial tokens (e.g., {\tt [MASK]} ) in the real downstream fine-tuning task.

To address this issue, we re-formalize punctuation 
restoration as Slot Tagging problem in fine-tuning stage and show an illustration in Figure \ref{fig:finetuning}. Specifically, given a non-punctuation input tokens $x=(x_1,x_2,x_3,x_4,...,x_n)$ and punctuation tags $y=(y_1,y_2,y_3,y_4,...,y_n)$, where $n$ denotes the length of the input tokens. We utilize Jieba\footnote{\url{https://github.com/fxsjy/jieba}} as our Chinese word segmentation (CWS) tool and convert the output tokens to original Chinese words. Then, the special token {\tt [M]} is inserted into each word.
\begin{equation}
    x_t=(x_1x_2,{\tt [M]},x_3x_4,{\tt [M]},...,x_n,{\tt [M]})
\end{equation}
where ${\tt [M]}$ is the abbreviated form of {\tt [MASK]}, $x_1x_2$ and $x_3x_4$ are Chinese words that combined with Chinese characters $x_1$,$x_2$ and $x_3$,$x_4$, respectively.

Then, we utilize a regular classifier to fill the ${\tt [M]}$ tokens, which consists of a feed-forward layer that is initialized by PMP Head. This classifier need to output a sequence $(\hat{y_1},\hat{y_2},...,\hat{y_K})$ in punctuation mark set $Y$, where $K$ is the number of the ${\tt [M]}$ tokens, and project final outputs as $ \hat{y} = (\textit{O}, \hat{y_1},\textit{O},\hat{y_2},...,\hat{y_K})$. The predicted punctuation marks of $x_1$ and $x_3$ are \textit{O}, in order not to insert punctuation marks into Chinese whole words.

\section{Corpus and Experimental Setup}
\label{sec:corpus}
\subsection{Corpus}

Aiming at training our PMP model on Chinese medical domain, we crawl up a huge amount of patient reports on medical website\footnote{Medical Reports: \url{https://bingli.iiyi.com/}. Chinese medicine books: \url{https://www.zysj.com.cn/}. Chinese medical encyclopedia: \url{http://www.a-hospital.com/}}. Then, we remove all html tags, duplicate data and patient privacy. We split our corpus into pre-training dataset and fine-tuning dataset, and the fine-tuning dataset is further divided into train, dev, test set. Pre-training dataset consists of 50 million tokens without labels. For fine-tuning dataset, we focused on three punctuation marks: $\{\textit{COMMA}, \textit{PERIOD}, \textit{COLON}\}$. The reason why exclamation and question marks are not used is that these two punctuation marks appear less frequently in Chinese clinical report and are not needed in practical application scenarios. We remove punctuation marks in each sequence, marking the label according to whether there is punctuation mark after each token. The format of fine-tuning dataset is similar to Named Entity Recognition (NER) dataset, which contains many non-punctuation sequences with punctuation marks as labels. Our corpus is named as CMRPT (Chinese Medical Reports Punctuation Transcription). Detailed statistics are given in Table \ref{tab:dataset}.

\begin{table*}[t] 
    \caption{Results in terms of precision (P \%), recall (R \%), and F1-score (F1 \%) on CMRPT.}
    \label{tab:result}
    \begin{tabular}{ccccccccccccc}
    \toprule
    \multirow{2}{*}{\textbf{Model}}&
    \multicolumn{3}{c}{\textbf{Comma}}&\multicolumn{3}{c}{\textbf{Period}}&\multicolumn{3}{c}{\textbf{Colon}}&\multicolumn{3}{c}{\textbf{Overall}}  \\
    \cmidrule(lr){2-4} \cmidrule(lr){5-7} \cmidrule(lr){8-10} \cmidrule(lr){11-13}
    &P&R&F1&P&R&F1&P&R&F1&P&R&F1\\
    \midrule
        RoBERTa-wwm &94.35 &95.71 &95.03 &89.88 &92.15 &91.00 &88.38 &83.77 &86.00 &90.87 &90.54 &90.68 \\
        RBT-6&93.45&86.58&89.88&88.94&84.41&86.62&82.57&81.44&81.99&88.32&84.14&86.16\\
        MiniRBT &88.78&88.54&88.66&86.03&79.23&82.49&87.58&76.04&81.41&87.46&81.27&84.18 \\ 
        TinyBERT &88.79 &89.01 &88.9 &86.77 &79.45 &82.95 &86.52 &74.42 &79.99 &87.36 &80.96 &83.95 \\
    \midrule

       \textbf{PMP-H768}&93.90&95.41&94.65&91.67&90.82&91.24&87.03&80.15&83.43&90.87&88.79&89.77 \\
        \textbf{PMP-H256}&93.28&87.07&90.07&88.68&85.06&86.83&84.16&77.04&80.32&88.71&83.06&85.74   \\
        \textbf{PMP-H312}&93.35&86.52&89.81&89.52&85.11&87.26&82.81&77.94&80.29&88.56&83.19&85.78\\
    \bottomrule
    \end{tabular}
\end{table*}

\begin{table}[t]
    \centering
    \caption{Comparison of PMP with ALBERT.}
    \label{tab:albert}
    \begin{tabular}{cccc}
    \toprule
    Model & P&R&F1 \\
    \midrule
    ALBERT&  89.44&85.87&87.54\\
    PMP-H768&  90.87 & 88.79 &89.77  \\
    \bottomrule
    \end{tabular}
\end{table}

\begin{table}[t]
    \centering
    \caption{Ablation Study. The model is PMP-H256 pre-trained with 30K steps}
    \label{tab:ablation}
    \begin{tabular}{l|ccc}
    \toprule
    \textbf{Model}&P&R&F1\\
    \midrule
    \textbf{PMP-H256}& 89.01 & 84.56 &86.72 \\
    w/o CE & 88.09 & 80.79 &84.15 \\
    w/o SCL& 88.09 & 83.19 &85.56 \\
    w/o KD & 87.91 & 89.05 &88.44 \\
    \bottomrule
    \end{tabular}
\end{table}

\subsection{Experimental Setup}
We employ RoBERTa-wwm~\cite{cui2021pre} as the teacher model. To compare with various distilled version of BERT, including RBT-6\footnote{\url{https://huggingface.co/hfl/rbt6}} (re-trained 6-layer RoBERTa-wwm model), MiniRBT~\cite{yao2023minirbt}, TinyBERT~\cite{jiao2020tinybert}, we present three students with similar hidden size and feed-forward size. The model structures are shown in Table \ref{tab:structure}.
In the pre-training stage, we set the maximum sequence length to 512 with a batch size of 256, and learning rate is 4e-4. For supervised contrastive learning, $\lambda$ and $\tau$ are 0.1 and 0.07 respectively. For knowledge distillation, the temperature is set to 8 and the number of training steps is 100K. In the fine-tuning stage, we use the AdamW~\cite{loshchilov2018decoupled} as optimizer with linear scheduler, a learning rate of 3e-5 and 0.1 warm-up step. We set weight-decay coefficient as 1e-5 and maximum gradient norms as 1.0. We train the model for 10 epochs on the training set, and the development set is used to save the best checkpoint by epoch. The experimental results are evaluated on the test set. We do pre-training tasks on 4 NVIDIA 80GB A100 GPUs and fine-tune on single NVIDIA RTX 4090 GPU. We run experiments in the fine-tuning stage 10 times with 10 different seeds and report average results. All experiments are evaluated using precision, recall and F1-score over the 4 punctuation marks. Our code and data are available at \url{https://github.com/rickltt/punctuation_restoration}.

\section{Experiment}
\label{sec:exps}
\subsection{Main Results}

Table \ref{tab:result} presents the results of fine-tuning different distilled models for punctuation restoration. Compared with RoBERTa-wwm, PMP-256 and PMP-312 achieve its 95\% performance with almost 10\% of the model size. With the same number of parameters (PMP-H256 v.s. MiniRBT, PMP-H312 v.s. TinyBERT, PMP-H768 v.s. RBT-6), PMP models achieve better performance, demonstrating that the overall feasibility and effectiveness of our pre-training stage. Then, PMP-H312 and PMP-H256 achieve similar performance of RBT-6 even though it has 5-6 times (10.3M v.s. 59.7M, 11.4 v.s. 59.7M) more parameters, proving that narrow and deep models yield higher performance than wide and shallow models. These results suggest that the pre-training stage effectively reduces the gap in size between teacher (RoBERTa-wwm) and student models (PMP), thus allowing PMP models to maintain excellent performance even with small sizes. Overall, our PMP models achieve the best performance among all the other distilled models, and remain almost 95\% performance of the teacher model. 

To compare with the current state-of-the-art light version of BERT, Table \ref{tab:albert} shows the results of ALBERT~\cite{lan2019albert} and PMP-H768. Our model achieves improvements of 4.3\% precision, 4.3\% recall and 4.3\% F1. The superiority of PMP with the same number of parameters of ALBERT proves that our proposed model is effective. Our experiments demonstrate that PMP has notable performance advantages over other models with the same number of parameters.

The competitive performance of our proposed method can be attributed to several key factors. Firstly, we distill pre-trained models, leveraging the power of transfer learning to capture general linguistic knowledge. Secondly, we incorporate supervised contrastive learning, which enhances the model's ability to discern fine-grained differences between inputs, crucial for accurate punctuation restoration. Finally, we introduce a novel auxiliary pre-training task: Punctuation Mark Prediction. This task specifically targets punctuation restoration, allowing the model to gain a deeper understanding of punctuation patterns and their contextual usage. The combination of these three elements results in a model that is well-suited for punctuation restoration, explaining the superior performance of our method.
 
\subsection{Ablation Study}
To explore which part plays a key role, we perform the ablation study in Table \ref{tab:ablation}. First, the declining result of eliminating CE verifies that punctuation prediction is effective to pre-training. Second, we remove SCL in the pre-training and the precision, recall and F1-score decrease 0.9, 1.3 and 1.1 points, respectively. Then, we directly remove knowledge distillation but keep CE and SCL. The performance of F1 score have a little boost from 86.72\% to 88.44\%, but with large model size and longer inference time. The metrics on F1-score decreases 1.16-2.57 points when we drop out SCL and CE, respectively. The studies confirm that the design of PMP can boost the effectiveness of punctuation restoration.

\section{Conclusion}
\label{sec:con}

This paper focuses on Chinese medical punctuation restoration and leverages supervised contrastive learning and a novel auxiliary pre-training task to enhance this task. To train a fast and light model, we employ knowledge distillation in pre-training stage. Extensive experiments and analyses show that PMP has promising performance on punctuation restoration.

\section{Acknowledgements}
This work was in part supported by Shenzhen Science and Technology Program (No:JCYJ20210324135809025).



\bibliographystyle{IEEEtran}
\bibliography{reference}

\end{document}